# STIFFNESS ANALYSIS OF MULTI-CHAIN PARALLEL ROBOTIC SYSTEMS WITH LOADING

**Anatol Pashkevich**[1,2], **Alexandr Klimchik**[1,2],
**Damien Chablat**[1], **Philippe Wenger**[1]

[1]*Ecole des Mines de Nantes,
4 rue Alfred-Kastler, Nantes 44307, France*

[2]*Institut de Recherches en Communications et Cybernétique de Nantes,
UMR CNRS 6597, 1 rue de la No, 44321 Nantes, France*

*Abstract*: The paper presents a new stiffness modelling method for multi-chain parallel robotic manipulators with flexible links and compliant actuating joints. In contrast to other works, the method involves a FEA-based link stiffness evaluation and employs a new solution strategy of the kinetostatic equations, which allows computing the stiffness matrix for singular postures and to take into account influence of the external forces. The advantages of the developed technique are confirmed by application examples, which deal with stiffness analysis of a parallel manipulator of the Orthoglide family.

## 1. INTRODUCTION

In modern manufacturing systems, parallel manipulators have become more and more popular for a variety of technological processes, including high-accuracy positioning and high-speed machining [1, 2]. This growing attention is inspired by their essential advantages over serial manipulators, which have already reached the dynamic performance limits In contrast, parallel manipulators are claimed to offer better accuracy, lower mass/inertia properties, and higher structural rigidity (i.e. stiffness-to-mass ratio) [3].

These features are induced by their specific kinematic structure, which resists the error accumulation in kinematic chains and allows convenient actuators location close to the manipulator base. This makes them attractive for innovative robotic systems, but practical utilization of the potential benefits requires development of efficient stiffness analysis techniques, which satisfy the computational speed and accuracy requirements of relevant design procedures.

Generally, the stiffness analysis evaluates the effect of the applied external torques and forces on the compliant displacements of the end-effector. Numerically, this property is defined through the "stiffness matrix" $\mathbf{K}$, which gives the relation between the translational/rotational displacement and the static forces/torques causing this transition. As follows from mechanics, $\mathbf{K}$ is 6×6 semi-definite non-negative matrix, where structure may be non-diagonal to represent the coupling between the translation and rotation [4, 5]. Similar to other manipulator properties (kinematical, for instance), the stiffness essentially depends on the force/torque direction and on the manipulator configuration [6].

Several approaches exist for the computation of the stiffness matrix, such as the Finite Element Analysis (FEA), the matrix structural analysis (MSA), and the virtual joint method (VJM). The FEA method is proved to be the most accurate and reliable, since the links/joints are modeled with its true dimension and shape. Its accuracy is limited by the discretisation step only. However, because of high computational expenses required for the repeated re-meshing, this method is usually applied at the final design stage.

The MSA method incorporates the main ideas of the FEA but operates with rather large flexible elements (beams, arcs, cables, etc.). This obviously yields reduction of the computational expenses and, in some cases, allows even obtaining an analytical stiffness matrix. This method gives a reasonable trade-off between the accuracy and computational time, provided that link approximation by the beam elements is realistic. Because it involves rather high-dimensional matrix operations, it is not attractive for the parametric stiffness analysis.

Finally, the VJM method, which is also referred to as the "lumped modeling", is based on the expansion of the traditional rigid model by adding virtual joints, which describe the elastic deformations of the manipulator components (links, joints and actuators). This approach originates from the work of [7], who evaluated parallel manipulator stiffness taking into account only the actuators compliance. At present, there are a number of variations and simplifications of the VJM method, which differ in modelling assumptions and numerical techniques. Generally, the lumped modelling provides acceptable accuracy in short computational time. However, it is very hypothetic and operates with simplified stiffness models that are composed of one-dimensional springs that do not take into account the coupling between the rotational and translational deflections. Recent modification of this method allows to extend it to the over-constrained manipulator and to apply it at any workspace point, including the singular ones [8].

It should be stressed that the standard stiffness analysis focuses on the unloaded structures, for which there were proposed several efficient semi-analytical technignes [9-11]. However, for the loaded working modes, the stiffness analysis is still an open problem. Besides, with respect to this case, several authors introduced a concept of the asymmetric Cartesian stiffness matrix [12-14], but this concept was recently revised by Kövecses and Angeles [5].

This paper presents a new stiffness modelling method for the loaded parallel manipulators, which is based on a multidimensional lumped-parameter model that replaces the link flexibility by localized 6-dof virtual springs that describe both the linear/rotational deflections and the coupling between them. The spring stiffness parameters are evaluated using FEA modelling to ensure higher accuracy. In addition, it employs a new solution strategy of the kinetostatic equations, which allows computing the stiffness matrix for the overconstrained architectures, including the singular manipulator postures. This gives almost the same accuracy as FEA but with essentially lower computational effort because it eliminates the model re-meshing through the workspace.

## 2. PROBLEM OF STIFFNESS MODELLING

### 2.1 Manipulator Architecture

Let us consider a general n-dof parallel manipulator, which consists of a mobile platform connected to a fixed base by n identical kinematics chains. Each chain includes an actuated joint "Ac" (prismatic or rotational) followed by a "Foot" and a "Leg" with a number of passive joints "Ps" inside (Fig. 1). Generally, certain geometrical conditions are assumed to be satisfied with respect to the passive joints to eliminate the undesired platform rotations and to achieve stability of desired motions. Typical examples of such architectures include 3-PUU translational parallel kinematic machine [15], Delta parallel robot [16], Orthoglide parallel manipulator that implements the 3-PRPaR architecture with parallelogram-type legs and translational active joints [17]. Here R, P, U and Pa denote the revolute, prismatic, universal and parallelogram joints, respectively.

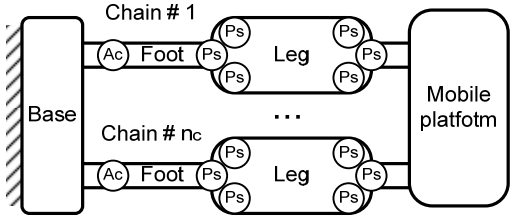

Fig. 1. Schematic diagram of a general *n*-dof parallel manipulator
(Ac – actuated joint, Ps – passive joints).

### 2.2 Basic Assumptions

To evaluate the manipulator stiffness, let us apply a modification of the virtual joint method (VJM), which is based on the lump modeling approach (Gosselin, 1990). According to this approach, the original rigid model should be extended by adding the virtual joints (localized springs), which describe elastic deformations of the links. Besides, virtual springs are included in the actuating joints to take into account stiffness of the control loop. . Under such assumptions, each kinematic chain of the manipulator can be described by a serial structure, which includes sequentially:

(a) a rigid link between the manipulator base and the ith actuating joint (part of the base platform) described by the constant homogenous transformation matrix $\mathbf{T}_{base}^{i}$;

(b) a 1-d.o.f. actuating joint with supplementary virtual spring, which is described by the homogenous matrix function $\mathbf{V}_{a}\left(q_{0}^{i}+\theta_{0}^{i}\right)$ where $q_{0}^{i}$ is the actuated coordinate and $\theta_{0}^{i}$ is the virtual spring coordinate;

(c) a rigid "Foot" linking the actuating joint and the leg, which is described by the constant homogenous transformation matrix $\mathbf{T}_{foot}$;

(d) a 6-d.o.f. virtual joint defining three translational and three rotational foot-springs, which are described by the homogenous matrix function $\mathbf{V}_{s}\left(\theta_{1}^{i},...,\theta_{6}^{i}\right)$, where $\{\theta_{1}^{i},\theta_{2}^{i},\theta_{3}^{i}\}$ and $\{\theta_{4}^{i},\theta_{5}^{i},\theta_{6}^{i}\}$ correspond to the elementary translations and rotations respectively;

(e) a 2-d.o.f. passive U-joint at the beginning of the leg allowing two independent rotations with angles $\{q_{1}^{i},q_{2}^{i}\}$, which is described by the homogenous matrix function $\mathbf{V}_{u1}\left(q_{1}^{i},q_{2}^{i}\right)$;

(f) a rigid "Leg" linking the foot to the movable platform, which is described by the constant homogenous matrix transformation $\mathbf{T}_{Leg}$;

(g) a 6-d.o.f. virtual joint defining three translational and three rotational leg-springs, which are described by the homogenous matrix function $\mathbf{V}_s\left(\theta_7^i, ..., \theta_{12}^i\right)$, where $\{\theta_7^i, \theta_8^i, \theta_9^i\}$ and $\{\theta_{10}^i, \theta_{11}^i, \theta_{12}^i\}$ correspond to the elementary translations and rotations, respectively;

(h) a 2-d.o.f. passive U-joint at the end of the leg allowing two independent rotations with angles $\{q_3^i, q_4^i\}$, which is described by the homogenous matrix function $\mathbf{V}_{u2}\left(q_3^i, q_4^i\right)$;

(i) a rigid link from the manipulator leg the end-effector (part of the movable platform) described by the homogenous matrix transformation $\mathbf{T}_{tool}^i$.

The expression defining the end-effector location subject to variations of all coordinates of a single kinematic chain may be written as follows

$$\mathbf{T}_i = \mathbf{T}_{base}^i \mathbf{V}_a\left(q_0^i + \theta_0^i\right) \mathbf{T}_{foot} \mathbf{V}_s\left(\theta_1^i, ..., \theta_6^i\right) \mathbf{V}_{u1}\left(q_1^i, q_2^i\right) \mathbf{T}_{Leg} \mathbf{V}_s\left(\theta_7^i, ..., \theta_{12}^i\right) \mathbf{V}_{u2}\left(q_3^i, q_4^i\right) \mathbf{T}_{tool}^i \quad (1)$$

where matrix function $\mathbf{V}_a(...)$ is either an elementary rotation or translation, matrix functions $\mathbf{V}_{u1}(...)$ and $\mathbf{V}_{u2}(...)$ are compositions of two successive rotations, and the spring matrix $\mathbf{V}_s(...)$ is composed of six elementary transformations.

## 2.3 Problem statement

In general, the stiffness model describes the resistance of an elastic body or a mechanism to deformations caused by an external force or torque [18]. For relatively small deformations, this property is defined through the "stiffness matrix" $\mathbf{K}$, which defines the linear relation

$$\mathbf{F} = \mathbf{K}(\mathbf{q_0}, \mathbf{\theta_0}) \cdot \mathbf{\Delta t} \qquad (2)$$

between the six-dimensional translational/rotational displacements $\mathbf{\Delta t} = (\Delta x, \Delta y, \Delta z, \Delta \varphi_x, \Delta \varphi_y, \Delta \varphi_z)^T$, and the static forces/torques $\mathbf{F} = (F_x, F_y, F_z, M_x, M_y, M_z)$ causing this transition. Here, the vector $\mathbf{q}_0 = (q_{01}, q_{02}, ..., q_{0n})^T$ includes all passive joint coordinates, the vector $\mathbf{\theta}_0 = (\theta_{01}, \theta_{02}, ..., \theta_{0m})^T$ collects all virtual joint coordinates, $n$ is the number of passive joins, $m$ is the number of virtual joints. Usually, the manipulator is assembled without internal preloading and the vector $\theta_0$ is equal to zero.

However, for the loaded mode, similar relation is defined in the neighborhood of the static equilibrium, which corresponds to another configuration of the manipulator $(\mathbf{q}, \mathbf{\theta})$, that is caused by external forces\torques $\mathbf{F}$. Respectively, in this case, the stiffness model describes the relation between the increments of the force $\mathbf{\delta F}$ and the position $\mathbf{\delta t}$

$$\mathbf{\delta F} = \mathbf{K}^F(\mathbf{q}, \mathbf{\theta}) \cdot \mathbf{\delta t} \qquad (3)$$

where $\mathbf{q} = \mathbf{q_0} + \mathbf{\Delta q}$ and $\mathbf{\theta} = \mathbf{\theta_0} + \mathbf{\Delta \theta}$ denote the loaded position of the manipulator, $\mathbf{\Delta q}$ and $\mathbf{\Delta \theta}$ are the deviations of the passive joint and virtual spring coordinates.

Let us also define the geometry of the manipulator in the Cartesian space as

$$\mathbf{t} = f(\mathbf{q}, \mathbf{\theta}), \qquad (4)$$

where the function $f(\mathbf{q}, \mathbf{\theta})$ is defined by the transformation (1), and the vector $\mathbf{t} = (\mathbf{p}, \mathbf{\varphi})^T$ describes the three-dimensional position $\mathbf{p} = (x, y, z)^T$ and orientation $\mathbf{\varphi} = (\varphi_x, \varphi_y, \varphi_z)^T$ of the end-effector with respect to the Cartesian axes.

Hence, the problem is to find the static equilibrium of the considered manipulator and to linearise relevant force/position relations.

## 3. STIFFNESS MODEL FOR THE LOADED MODE

To derive the desired stiffness model, let us divide the problem into three sequential subtasks that are solved for each kinematic chain separately: (i) computing the stiffness matrix for the unloaded mode, (ii) finding the static equilibrium for the loaded configuration, and (iii) obtaining the stiffness model for the loaded mode. At the final stage, these results for separate kinematic chains are aggregated, in order to obtain the stiffness of the entire manipulator.

### 3.1 Stiffness model in the neighborhood of unloaded configuration

Let us define the unloaded configuration as $\mathbf{t}_0 = f(\mathbf{q}_0, \mathbf{\theta}_0)$, where $\mathbf{q}_0$ is computed via the inverse kinematic and $\mathbf{\theta}_0$ is equal to zero (since there are no preloads in the springs). Let us also assume that the external force $\mathbf{F}$ relocates the manipulator to the position $\mathbf{t} = f(\mathbf{q}_0 + \Delta\mathbf{q}, \mathbf{\theta}_0 + \Delta\mathbf{\theta})$, which for small displacements may be expressed as

$$\mathbf{t} = \mathbf{t}_0 + \mathbf{J}_\theta \cdot \Delta\mathbf{\theta} + \mathbf{J}_q \cdot \Delta\mathbf{q} \tag{5}$$

where $\mathbf{J}_\theta = \left.\dfrac{\partial f(\mathbf{q}, \mathbf{\theta})}{\partial \mathbf{\theta}}\right|_{\substack{\mathbf{q}=\mathbf{q}_0 \\ \mathbf{\theta}=\mathbf{\theta}_0}}$ and $\mathbf{J}_q = \left.\dfrac{\partial f(\mathbf{q}, \mathbf{\theta})}{\partial \mathbf{q}}\right|_{\substack{\mathbf{q}=\mathbf{q}_0 \\ \mathbf{\theta}=\mathbf{\theta}_0}}$

are the kinematic Jacobians with respect to the coordinates θ, **q**, which may be computed from (1) analytically or semi-analytically, using the factorization technique proposed in [11].

For the kinetostatic model, which describes the force-and-motion relation, it is necessary to introduce additional equations that define the virtual joint reactions to the corresponding spring deformations. For analytical convenience, all relevant expressions may be collected in a single matrix equation

$$\boldsymbol{\tau}_{\boldsymbol{\theta}} = \mathbf{K}_{\boldsymbol{\theta}} \cdot \boldsymbol{\theta} \qquad (6)$$

where $\boldsymbol{\tau}_{\boldsymbol{\theta}} = \left(\tau_{\theta,1},\ \tau_{\theta,2},\ ...,\ \tau_{\theta,m}\right)^{T}$ is the aggregated vector of the virtual joint reactions, $\mathbf{K}_{\boldsymbol{\theta}} = diag\left(\mathbf{K}_{\boldsymbol{\theta},1},\ \mathbf{K}_{\boldsymbol{\theta},2},\ ...,\ \mathbf{K}_{\boldsymbol{\theta},m}\right)$ is the aggregated spring stiffness matrix of the size m×m, and $\mathbf{K}_{\boldsymbol{\theta},i}$ is the spring stiffness matrix of the corresponding link. Similarly, one can define the aggregated vector of the passive joint reactions $\boldsymbol{\tau}_{\mathbf{q}} = \left(\tau_{q,1},\ \tau_{q,2},\ ...,\ \tau_{q,n}\right)^{T}$ but, at the equilibrium, all its components must be equal to zero

$$\boldsymbol{\tau}_{\mathbf{q}} = \mathbf{0} \qquad (7)$$

Further, let us apply the principle of virtual work assuming that the joints are given small, arbitrary virtual displacements $\Delta\boldsymbol{\theta}$ in the equilibrium neighborhood. Then, the virtual work of the external force **F** applied to the end-effector along the corresponding displacement $\Delta\mathbf{t} = \mathbf{J}_{\boldsymbol{\theta}} \cdot \Delta\boldsymbol{\theta} + \mathbf{J}_{\mathbf{q}} \cdot \Delta\mathbf{q}$ is equal to the sum $\left(\mathbf{F}^{T}\mathbf{J}_{\boldsymbol{\theta}}\right) \cdot \Delta\boldsymbol{\theta} + \left(\mathbf{F}^{T}\mathbf{J}_{\mathbf{q}}\right) \cdot \Delta\mathbf{q}$. For the internal forces, the virtual work includes only one component $-\boldsymbol{\tau}_{\boldsymbol{\theta}}^{T} \cdot \Delta\boldsymbol{\theta}$, since the passive joints do not produce the force/torque reactions (the minus sign takes into account the adopted directions for the virtual spring forces/torques). Therefore, since in the static equilibrium the total virtual work is equal to zero for any virtual displacement, the equilibrium conditions may be written as

$$\mathbf{J}_\theta^T \cdot \mathbf{F} = \boldsymbol{\tau}_\theta; \quad \mathbf{J}_q^T \cdot \mathbf{F} = \mathbf{0} \tag{8}$$

This gives additional expressions describing the force/torque propagation from the joints to the end-effector.

Hence, the complete kinetostatic model consists of four matrix equations (5)…(8) where either $\mathbf{F}$ or $\Delta \mathbf{t}$ are treated as known, and the remaining variables are considered as unknowns. Since the matrix $\mathbf{K}_\theta$ is non-singular (it describes the stiffness of the virtual sprigs), the variables $\Delta \boldsymbol{\theta}$ can be expressed via $\mathbf{F}$ using equations (5)…(8). This yields substitution $\Delta \boldsymbol{\theta} = \mathbf{K}_\theta^{-1} \cdot \mathbf{J}_\theta^T \cdot \mathbf{F}$ allowing reducing the kinetostatic model to system of two matrix equations with unknowns $\mathbf{F}$ and $\Delta \mathbf{q}$, which can be written in the matrix form as

$$\begin{bmatrix} \mathbf{S}_\theta & \mathbf{J}_q \\ \mathbf{J}_q^T & \mathbf{0} \end{bmatrix} \cdot \begin{bmatrix} \mathbf{F} \\ \Delta \mathbf{q} \end{bmatrix} = \begin{bmatrix} \Delta \mathbf{t} \\ \mathbf{0} \end{bmatrix} \tag{9}$$

where the sub-matrix $\mathbf{S}_\theta = \mathbf{J}_\theta \cdot \mathbf{K}_\theta^{-1} \cdot \mathbf{J}_\theta^T$ describes the spring compliance relative to the end-effector, and the sub-matrix $\mathbf{J}_q$ takes into account the passive joint influence on the end-effector motions. Therefore, for a separate kinematic chain, the desired stiffness matrix $\mathbf{K}$ defining the motion-to-force mapping

$$\mathbf{F} = \mathbf{K} \cdot \Delta \mathbf{t}, \tag{10}$$

can be computed by the direct inversion of (6+n)×(6+n) matrix in the left-hand side of (10) and extracting from it the 6×6 sub-matrix with indices corresponding to $\mathbf{S}_\theta$.

### 3.2 Static equilibrium for the loaded configuration

Let us assume that, due to the external force **F**, the manipulators is relocated from the initial (unloaded) position $\mathbf{t}_0 = f(\mathbf{q}_0, \mathbf{\theta}_0)$ to a new position $\mathbf{t} = f(\mathbf{q}, \mathbf{\theta})$, which satisfies the condition of the mechanical equilibrium. If the displacement $\Delta \mathbf{t} = \mathbf{t} - \mathbf{t}_0$ is rather small, the new configuration $(\mathbf{q}, \mathbf{\theta})$ can be computed easily, using results from the previous subsection. However, in general case, the stiffness model is highly non-linear and computing $(\mathbf{q}, \mathbf{\theta})$ requires some additional efforts. Besides, for computational reasons, let us consider the dual problem that deals with determining the external force **F** and the manipulator configuration $(\mathbf{q}, \mathbf{\theta})$ that correspond to the output position $\mathbf{t}$.

For the considered problem, the basic equations can be written as

$$\mathbf{t} = f(\mathbf{q}, \mathbf{\theta}); \quad \mathbf{J}_\theta^T(\mathbf{q}, \mathbf{\theta}) \cdot \mathbf{F} = \mathbf{K}_\theta \cdot \mathbf{\theta}; \quad \mathbf{J}_q^T(\mathbf{q}, \mathbf{\theta}) \cdot \mathbf{F} = \mathbf{0}, \qquad (11)$$

where the first equation defines the manipulator geometry and the remaining ones are derived from statics. It is evident that there is no general method for analytical solution of this system and it is required to apply numerical techniques.

To derive the numerical algorithm, let us linearise the kinematic equation in the neighborhood of the $(\mathbf{q}_i, \mathbf{\theta}_i)$

$$\mathbf{t} = f(\mathbf{q}_i, \mathbf{\theta}_i) + \mathbf{J}_q(\mathbf{q}_i, \mathbf{\theta}_i) \cdot (\mathbf{q}_{i+1} - \mathbf{q}_i) + \mathbf{J}_\theta(\mathbf{q}_i, \mathbf{\theta}_i) \cdot (\mathbf{\theta}_{i+1} - \mathbf{\theta}_i) \qquad (12)$$

and rewrite the static equations as

$$\mathbf{J}_\theta^T(\mathbf{q}_i, \mathbf{\theta}_i) \cdot \mathbf{F}_{i+1} = \mathbf{K}_\theta \, \mathbf{\theta}_{i+1}; \quad \mathbf{J}_q^T(\mathbf{q}_i, \mathbf{\theta}_i) \cdot \mathbf{F}_{i+1} = \mathbf{0} \qquad (13)$$

This leads to a linear algebraic system of equations with respect to $(\mathbf{q}_{i+1}, \mathbf{\theta}_{i+1}, \mathbf{F}_{i+1})$

$$\begin{bmatrix} \mathbf{J}_q(\mathbf{q}_i,\boldsymbol{\theta}_i) & \mathbf{J}_\theta(\mathbf{q}_i,\boldsymbol{\theta}_i) & \mathbf{0} \\ \mathbf{0} & -\mathbf{K}_\theta & \mathbf{J}_\theta^T(\mathbf{q}_i,\boldsymbol{\theta}_i) \\ \mathbf{0} & \mathbf{0} & \mathbf{J}_q^T(\mathbf{q}_i,\boldsymbol{\theta}_i) \end{bmatrix} \cdot \begin{bmatrix} \mathbf{q}_{i+1} \\ \boldsymbol{\theta}_{i+1} \\ \mathbf{F}_{i+1} \end{bmatrix} = \begin{bmatrix} \mathbf{t} - \mathbf{f}(\mathbf{q}_i,\boldsymbol{\theta}_i) - \mathbf{J}_q(\mathbf{q}_i,\boldsymbol{\theta}_i)\cdot\mathbf{q}_i - \mathbf{J}_\theta(\mathbf{q}_i,\boldsymbol{\theta}_i)\cdot\boldsymbol{\theta}_i \\ \mathbf{0} \\ \mathbf{0} \end{bmatrix} \quad (14)$$

which gives the following iterative scheme

$$\begin{bmatrix} \mathbf{F}_{i+1} \\ \mathbf{q}_{i+1} \end{bmatrix} = \begin{bmatrix} \mathbf{J}_\theta(\mathbf{q}_i,\boldsymbol{\theta}_i)\cdot\mathbf{K}_\theta^{-1}\cdot\mathbf{J}_\theta^T(\mathbf{q}_i,\boldsymbol{\theta}_i) & \mathbf{J}_q(\mathbf{q}_i,\boldsymbol{\theta}_i) \\ \mathbf{J}_q^T(\mathbf{q}_i,\boldsymbol{\theta}_i) & \mathbf{0} \end{bmatrix}^{-1} \begin{bmatrix} \mathbf{t} - \mathbf{f}(\mathbf{q}_i,\boldsymbol{\theta}_i) - \mathbf{J}_q(\mathbf{q}_i,\boldsymbol{\theta}_i)\cdot\mathbf{q}_i - \mathbf{J}_\theta(\mathbf{q}_i,\boldsymbol{\theta}_i)\cdot\boldsymbol{\theta}_i \\ \mathbf{0} \end{bmatrix}$$

$$\boldsymbol{\theta}_{i+1} = \mathbf{K}_\theta^{-1}\cdot\mathbf{J}_\theta^T(\mathbf{q}_i,\boldsymbol{\theta}_i)\cdot\mathbf{F}_{i+1}$$

(15)

where the starting point can be chosen using the non-loaded configuration, i.e. $(\mathbf{q}_0,\boldsymbol{\theta}_0)$.

As follows from computational experiments, for typical values of deformations the proposed iterative algorithm possesses rather good convergence (3-5) iterations are usually enough). However, in the case of buckling or in the area of multiple equilibriums, the problem of convergence becomes rather critical and highly depends on the initial guess. Further enhancement of this algorithm may be based on the full-scale Newton-Raphson technique (i.e. linearization of the static equations in addition to the kinematic one), this obviously increases computational expenses but potentially improves convergence.

**3.3 Stiffness model for the loaded configuration**

In the neighborhood of the loaded configurations, the stiffness model is defined with respect to the force and position increments $\delta\mathbf{F}$, $\delta\mathbf{t}$, which are assumed to be small (see equation (3)). To derive this model, let us consider two equilibriums corresponding to the manipulator variables $(\mathbf{F},\mathbf{q},\boldsymbol{\theta},\mathbf{t})$ and $(\mathbf{F}+\delta\mathbf{F},\mathbf{q}+\delta\mathbf{q},\boldsymbol{\theta}+\delta\boldsymbol{\theta},\mathbf{t}+\delta\mathbf{t})$ respectively.

For this settings, the kinematic equation is reduced to

$$\delta \mathbf{t} = \mathbf{J}_\theta(\mathbf{q},\,\theta) \cdot \delta\theta + \mathbf{J}_q(\mathbf{q},\,\theta) \cdot \delta\mathbf{q}, \tag{16}$$

while the statics yields two set of equations

$$\mathbf{J}_\theta^T(\mathbf{q},\,\theta) \cdot \mathbf{F} = \mathbf{K}_\theta \cdot \theta; \qquad \mathbf{J}_q^T(\mathbf{q},\,\theta) \cdot \mathbf{F} = \mathbf{0} \tag{17}$$

and

$$(\mathbf{F} + \delta\mathbf{F}) \cdot (\mathbf{J}_\theta + \delta\mathbf{J}_\theta)^T = \mathbf{K}_\theta \cdot (\theta + \delta\theta); \qquad (\mathbf{F} + \delta\mathbf{F}) \cdot (\mathbf{J}_q + \delta\mathbf{J}_q)^T = \mathbf{0} \tag{18}$$

where $\delta\mathbf{J}_q(\mathbf{q},\theta)$ and $\delta\mathbf{J}_\theta(\mathbf{q},\theta)$ are the differentials of the Jacobians due to changes in $(\mathbf{q},\,\theta)$. After relevant transformation and neglecting high-order small terms, equations (17), (18) may be rewritten as

$$\begin{aligned}\mathbf{J}_\theta^T(\mathbf{q},\theta) \cdot \delta\mathbf{F} + \mathbf{H}_{\theta q}^F(\mathbf{q},\theta) \cdot \delta\mathbf{q} + \mathbf{H}_{\theta\theta}^F(\mathbf{q},\theta) \cdot \delta\theta &= \mathbf{K}_\theta \cdot \delta\theta \\ \mathbf{J}_q^T(\mathbf{q},\theta) \cdot \delta\mathbf{F} + \mathbf{H}_{qq}^F(\mathbf{q},\theta) \cdot \delta\mathbf{q} + \mathbf{H}_{q\theta}^F(\mathbf{q},\theta) \cdot \delta\theta &= \mathbf{0}\end{aligned} \tag{19}$$

where $\mathbf{H}_{qq}^F,\ \mathbf{H}_{q\theta}^F,\ \mathbf{H}_{\theta q}^F,\ \mathbf{H}_{\theta\theta}^F$, are the Hessian matrices of the scalar function $\mathbf{F}^T \cdot f(\mathbf{q},\theta)$:

$$\begin{aligned}\mathbf{H}_{qq}^F &= \frac{\partial^2}{\partial \mathbf{q}^2}\big(\mathbf{F}^T \cdot f(\mathbf{q},\,\theta)\big); & \mathbf{H}_{q\theta}^F &= \frac{\partial^2}{\partial \mathbf{q} \partial \theta}\big(\mathbf{F}^T \cdot f(\mathbf{q},\,\theta)\big); \\ \mathbf{H}_{\theta q}^F &= \frac{\partial^2}{\partial \theta \partial \mathbf{q}}\big(\mathbf{F}^T \cdot f(\mathbf{q},\,\theta)\big); & \mathbf{H}_{\theta\theta}^F &= \frac{\partial^2}{\partial \theta^2}\big(\mathbf{F}^T \cdot f(\mathbf{q},\,\theta)\big)\end{aligned} \tag{20}$$

This allows to apply substitution for $\delta\theta$ and to obtain system of two matrix equations with unknowns $\delta\mathbf{F}$ and $\delta\mathbf{q}$

$$\begin{bmatrix}\mathbf{J}_\theta \cdot \mathbf{k}_\theta^F \cdot \mathbf{J}_\theta^T & \mathbf{J}_q + \mathbf{J}_\theta \cdot \mathbf{k}_\theta^F \cdot \mathbf{K}_{\theta q}^F \\ \mathbf{J}_q^T + \mathbf{K}_{q\theta}^F \cdot \mathbf{k}_\theta^F \cdot \mathbf{J}_\theta^T & \mathbf{K}_{qq}^F + \mathbf{K}_{q\theta}^F \cdot \mathbf{k}_\theta^F \cdot \mathbf{K}_{\theta q}^F\end{bmatrix} \cdot \begin{bmatrix}\delta\mathbf{F} \\ \delta\mathbf{q}\end{bmatrix} = \begin{bmatrix}\delta\mathbf{t} \\ \mathbf{0}\end{bmatrix}, \tag{21}$$

which generalizes (9) for the case of the loaded equilibrium. Here $\mathbf{k}_\theta^F = \big(\mathbf{K}_\theta - \mathbf{K}_{\theta\theta}^F\big)^{-1}$.

Therefore, for a separate kinematic chain, the desired stiffness matrix $\mathbf{K}^F$ defining the displacement-to-force mapping (3) can be computed by direct inversion of the matrix in the left-hand side of (21) and extracting from it the left-upper 6×6 sub-matrix. Finally, when the stiffness matrices for all kinematic chains are computed, the stiffness of the entire multi chain manipulator can be found by simple summation $\mathbf{K}^F = \sum_{i=1}^{n} \mathbf{K}_i^F$. This follows from the superposition principle, since the total external force corresponding to the end-effector displacement $\delta\mathbf{t}$ (the same for all kinematic chains) can be expressed as $\mathbf{F} = \sum_{i=1}^{3} \mathbf{F}_i$ where $\mathbf{F}_i = \mathbf{K}_i^F \cdot \delta\mathbf{t}$. It should be stressed that usually the matrices $\mathbf{K}_i^F$ are not invertible but for the entire manipulator, the stiffness matrix $\mathbf{K}^F = \sum_{i=1}^{3} \mathbf{K}_i^F$ is positive definite and invertible for all non-singular postures.

## 4. EVALUATING THE MODEL PARAMETERS

### 4.1 Actuator compliance

The actuator compliance, describing by the scalar parameter $\mathbf{K}_{ctr}$ and by 6×6 matrix $\mathbf{K}_{act}$, depends on both the servomechanism mechanics and the control algorithm. Since most modern actuators implement the digital PID control, the main contribution to the compliance is produces by the mechanical transmissions. The latter are usually located outside the feedback-control loop and consist of screws, gears, shafts, belts, etc., whose flexibility is comparable with the flexibility of the manipulator links. Because of the complicated mechanical structure of the servomechanisms, these parameters are usually evaluated from static load experiments, by applying the linear regression to the experimental data.

## 4.2 Link compliance

Following a general methodology, the compliance of a manipulator link (foots and legs) is described by 6×6 symmetrical positive definite matrices $\mathbf{K}_{Foot}, \mathbf{K}_{Leg}$ corresponding to 6-d.o.f. springs with relevant coupling between translational and rotational deformations. This distinguishes our approach from other lumped-based techniques, where the coupling is neglected and only a subset of deformations is taken into account (presented by a set of 1-d.o.f. springs).

The simplest way to obtain these matrices is to approximate the link by a beam element for which the non-zero elements of the compliance matrix may be expressed analytically. However, for certain link geometries, the accuracy of a single-beam approximation can be insufficient. In this case the link can be approximated by a serial chain of the beams, whose compliance is evaluated by applying the same method (i.e. considering the kinematic chain with 6-d.o.f. virtual springs, but without passive joints). This leads to the resulting compliance matrix $\mathbf{K}_{Link}^{-1} = \mathbf{J}_b \cdot \mathbf{K}_b^{-1} \cdot \mathbf{J}_b^T$, where $\mathbf{J}_b$ and $\mathbf{K}_b^{-1}$ incorporate the Jacobian and the compliance matrices for all virtual springs.

## 4.3 FEA-based evaluation of model parameters

For complex link geometries, the most reliable results can be obtained from the FEA modeling. To apply this approach, the CAD model of each link should be extended by introducing an auxiliary 3D object, a "pseudo-rigid" body, which is used as a reference for the compliance evaluation. Besides, the link origin must be fixed relative to the global coordinate system. Then, sequentially and separately applying forces $F_x, F_y, F_z$ and torques $M_x, M_y, M_z$ to the reference object, it is possible to evaluate corresponding linear and angular displacements, which allow computing the stiffness matrix columns. The main difficulty here is to obtain accurate displacement values by using proper FEA-discretization ("mesh size"). As follows from our study, the single-beam approximation of the Orthoglide links gives accuracy of about 50%, and the four-beam approximation improves it up to 30% only (compared to the FEA-based method that is proved producing very accurate results).

It worth mentioning that here, in contrast to the straightforward FEA-modeling, which requires re-computing for each manipulator posture, it is needed only a single evaluation of the link stiffness. The latter essentially improves the computational speed.

## 5. APPLICATION EXAMPLES

To demonstrate efficiency of the proposed methodology, let us apply it to the comparative stiffness analysis of two 3-d.o.f. translational mechanism, which employ Orthoglide architecture. CAD models of these mechanisms are presented in Fig. 2.

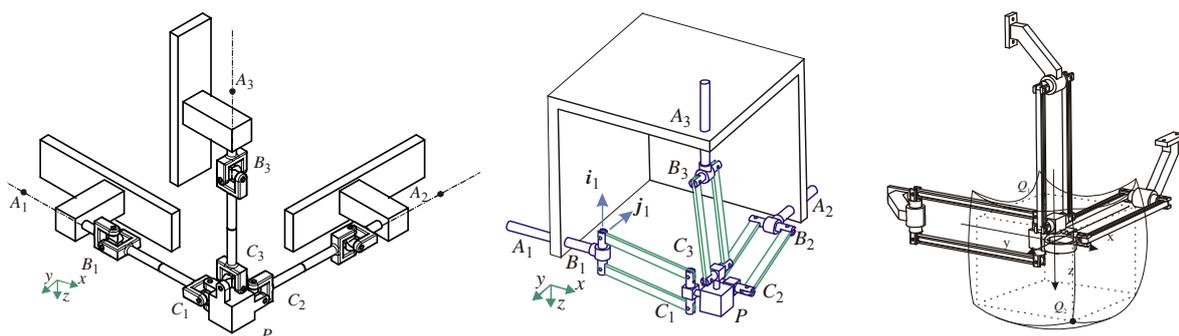

Fig. 2. Kinematics of two 3-dof translational mechanisms employing the Orthoglide architecture

First, let us derive the stiffness model for the simplified Orthoglide mechanics (3-PUU), where the legs are comprised of equivalent limbs with U-joints at the ends. Accordingly, to retain major compliance properties, the limb geometry corresponds to the parallelogram bars with doubled cross-section area. The geometrical models of separate kinematic chains can be described by the expression (1), where the product components are defined via the standard translational/rotational operators. Because for the rigid manipulator the end-effector moves with only translational motions, the nominal values of the passive joint coordinates are subject to the specific constrains $q_3 = -q_2$, $q_4 = -q_1$, which are implicitly incorporated in the direct/inverse kinematics.

For the second architecture (3-PRPrP) it is necessary to derive first the stiffness matrix of the parallelogram. Using the adopted notations, the parallelogram equivalent model may be written as

$$\mathbf{T}_{Plg} = \mathbf{R}_y(q_2) \cdot \mathbf{T}_x(L) \cdot \mathbf{R}_y(-q_2) \cdot \mathbf{V}_s(\theta_7, \mathrm{K}\, \theta_{12}) \tag{22}$$

where, compared to the above case, the third passive joint is eliminated (it is implicitly assumed that $q_3 = -q_2$). On the other hand, the original parallelogram may be split into two serial kinematic chains (the "upper" and "lower" ones). Hence, the parallelogram compliance matrix may be also derived using the proposed technique that yields an analytical expression [11].

Using this model and applying the proposed technique, there were computed the compliance matrices for both architectures and for three typical manipulator postures $Q_0$, $Q_1$ or $Q_2$ (see Tables 1, 2). As follows from the comparison, the parallelograms allow increasing the rotational stiffness roughly in 10 times.

The second conclusion is related to the stiffness comparison for the unloaded and loaded modes. It was assumed that the loading (Table 3) leads to the translational deflection of 0.5 mm in all Cartesian directions but the platform orientation remains the same. The obtained results confirm influence of the loading on the manipulator stiffness. In particular, some elements of the stiffness matrix may increase up to 45%, depending on the working point ($Q_0$, $Q_1$ or $Q_2$). Also, the 3-PUU manipulator is more sensitive to the external loading than its counterpart 3-PRPaR. This justifies application of 3-PRPaR architecture for high-speed machining.

Table 1: Translational and rotational stiffness of the 3-PUU manipulator
(unloaded and loaded modes)

| | Point $Q_0$ x,y,z = 0.00 mm | Point $Q_1$ x,y,z = -73.65 mm | Point $Q_2$ x,y,z = 126.35 mm |
|---|---|---|---|
| Manipulator position | 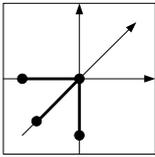 | 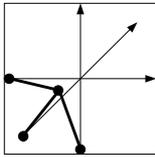 | 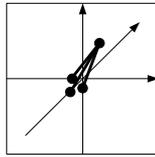 |
| | Unloaded mode | | |
| $k_{tran} \cdot 10^4$ [mm/N] | $\begin{bmatrix} 2.78 & 0 & 0 \\ 0 & 2.78 & 0 \\ 0 & 0 & 2.78 \end{bmatrix}$ | $\begin{bmatrix} 10.9 & 5.5 & 5.5 \\ 5.5 & 10.9 & 5.5 \\ 5.5 & 5.5 & 10.9 \end{bmatrix}$ | $\begin{bmatrix} 71.3 & -35.0 & -35.0 \\ -35.0 & 71.3 & -35.0 \\ -35.0 & -35.0 & 71.3 \end{bmatrix}$ |
| $k_{rot} \cdot 10^7$ [rad/N·mm] | $\begin{bmatrix} 20.9 & 0 & 0 \\ 0 & 20.9 & 0 \\ 0 & 0 & 20.9 \end{bmatrix}$ | $\begin{bmatrix} 24.1 & 7.5 & 7.5 \\ 7.5 & 24.1 & 7.5 \\ 7.5 & 7.5 & 24.1 \end{bmatrix}$ | $\begin{bmatrix} 25.8 & -7.4 & -7.4 \\ -7.4 & 25.8 & -7.4 \\ -7.4 & -7.4 & 25.8 \end{bmatrix}$ |
| | Loaded mode, $\Delta \mathbf{t}$ = ( 0.5, 0.5, 0.5, 0, 0, 0) | | |
| $k_{tran} \cdot 10^4$ [mm/N] | $\begin{bmatrix} 2.74 & -0.02 & -0.02 \\ -0.02 & 2.74 & -0.02 \\ -0.02 & -0.02 & 2.74 \end{bmatrix}$ | $\begin{bmatrix} 10.5 & 5.3 & 5.3 \\ 5.3 & 10.5 & 5.3 \\ 5.3 & 5.3 & 10.5 \end{bmatrix}$ | $\begin{bmatrix} 39.1 & -18.9 & -18.9 \\ -18.9 & 39.1 & -18.9 \\ -18.9 & -18.9 & 39.1 \end{bmatrix}$ |
| $k_{rot} \cdot 10^7$ [rad/N·mm] | $\begin{bmatrix} 16.7 & -0.02 & -0.02 \\ -0.02 & 16.7 & -0.02 \\ -0.02 & -0.02 & 16.7 \end{bmatrix}$ | $\begin{bmatrix} 22.0 & 6.0 & 6.0 \\ 6.0 & 22.0 & 6.0 \\ 6.0 & 6.0 & 22.0 \end{bmatrix}$ | $\begin{bmatrix} 15.4 & -0.7 & -0.7 \\ -0.7 & 15.4 & -0.7 \\ -0.7 & -0.7 & 15.4 \end{bmatrix}$ |

Table 2: Translational and rotational stiffness of the 3-PRPaR manipulator
(unloaded and loaded modes)

| Manipulator position | Point $Q_0$ x,y,z = 0.00 mm 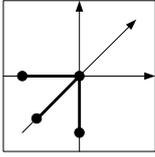 | Point $Q_1$ x,y,z = -73.65 mm 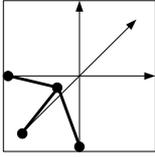 | Point $Q_2$ x,y,z = 126.35 mm 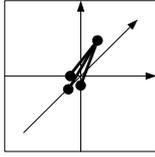 |
|---|---|---|---|
| Unloaded mode ||||
| $k_{tran} \cdot 10^4$ [mm/N] | $\begin{bmatrix} 2.78 & 0 & 0 \\ 0 & 2.78 & 0 \\ 0 & 0 & 2.78 \end{bmatrix}$ | $\begin{bmatrix} 9.86 & 5.80 & 5.80 \\ 5.80 & 9.86 & 5.80 \\ 5.80 & 5.80 & 9.86 \end{bmatrix}$ | $\begin{bmatrix} 21.2 & -10.2 & -10.2 \\ -10.2 & 21.2 & -10.2 \\ -10.2 & -10.2 & 21.2 \end{bmatrix}$ |
| $k_{rot} \cdot 10^7$ [rad/N·mm] | $\begin{bmatrix} 1.94 & 0 & 0 \\ 0 & 1.94 & 0 \\ 0 & 0 & 1.94 \end{bmatrix}$ | $\begin{bmatrix} 2.06 & -0.32 & -0.32 \\ -0.32 & 2.06 & -0.32 \\ -0.32 & -0.32 & 2.06 \end{bmatrix}$ | $\begin{bmatrix} 2.65 & 1.14 & 1.14 \\ 1.14 & 2.65 & 1.14 \\ 1.14 & 1.14 & 2.65 \end{bmatrix}$ |
| Loaded mode, $\Delta \mathbf{t} = (0.5, 0.5, 0.5, 0, 0, 0)$ ||||
| $k_{tran} \cdot 10^4$ [mm/N] | $\begin{bmatrix} 2.74 & -0.02 & -0.02 \\ -0.02 & 2.74 & -0.02 \\ -0.02 & -0.02 & 2.74 \end{bmatrix}$ | $\begin{bmatrix} 9.55 & 5.52 & 5.52 \\ 5.52 & 9.55 & 5.52 \\ 5.52 & 5.52 & 9.55 \end{bmatrix}$ | $\begin{bmatrix} 16.7 & -7.9 & -7.9 \\ -7.9 & 16.7 & -7.9 \\ -7.9 & -7.9 & 16.7 \end{bmatrix}$ |
| $k_{rot} \cdot 10^7$ [rad/N·mm] | $\begin{bmatrix} 1.88 & 0.01 & 0.01 \\ 0.01 & 1.88 & 0.01 \\ 0.01 & 0.01 & 1.88 \end{bmatrix}$ | $\begin{bmatrix} 2.05 & -0.31 & -0.31 \\ -0.31 & 2.05 & -0.31 \\ -0.31 & -0.31 & 2.05 \end{bmatrix}$ | $\begin{bmatrix} 2.50 & 1.28 & 1.28 \\ 1.28 & 2.50 & 1.28 \\ 1.28 & 1.28 & 2.50 \end{bmatrix}$ |

Table 3: Wrenches for the loaded mode( $\Delta \mathbf{t}$ = ( 0.5, 0.5, 0.5, 0, 0, 0) )

| Manipulator architecture | Point $Q_0$ x,y,z = 0.00 mm | Point $Q_1$ x,y,z = -73.65 mm | Point $Q_2$ x,y,z = 126.35 mm |
|---|---|---|---|
| 3-PUU | $F = 1823\ N$ $M = -101\ N \cdot mm$ | $F = 234\ N$ $M = 524\ N \cdot mm$ | $F = 4104\ N$ $M = -2525\ N \cdot mm$ |
| 3-PRPaR | $F = 1823\ N$ $M = 63.54\ N \cdot mm$ | $F = 248\ N$ $M = -6045\ N \cdot mm$ | $F = 9418\ N$ $M = 68376\ N \cdot mm$ |

## 6. CONCLUSIONS

The paper proposes a new systematic method for computing the stiffness matrix of multi-chain parallel robotic manipulators in the presence of the external loading applied to the end-platform. It is based on multidimensional lumped model of the flexible links, whose parameters are evaluated via the FEA modeling and describe both the translational/rotational compliances and the coupling between them. In contrast to previous works, the method employs a new solution strategy of the kinetostatic equations and allows computing the stiffness matrices for any given manipulator posture, including the singular ones.

The efficiency of the proposed method was demonstrated through application examples, which deal with comparative stiffness analysis of two parallel manipulators of the Orthoglide family. Relevant simulation results have confirmed essential advantages of the parallelogram-based architecture and validated adopted design of the Orthoglide prototype. In future work, the method will be extended to other parallel architectures composed of several identical kinematic chains and for other types of external loading.

## 7  ACKNOWLEDGEMENTS

The work presented in this paper was partially funded by the Region "Pays de la Loire", France and by the EU commission (project NEXT).